\documentclass{article}

\usepackage[preprint]{neurips_2023}

\usepackage[utf8]{inputenc} 
\usepackage[T1]{fontenc}    
\usepackage{hyperref}       
\usepackage{url}            
\usepackage{booktabs}       
\usepackage{amsfonts}       
\usepackage{nicefrac}       
\usepackage{microtype}      
\usepackage{xcolor}         
\usepackage{blindtext}      
\usepackage{graphicx}
\usepackage{color,colortbl}
\definecolor{cycle2}{RGB}{55, 126, 184}
\usepackage{bm}
\usepackage{amsmath}
\usepackage{bm}
\usepackage{pgfplots}
\pgfplotsset{compat=1.17} 
\usepackage{comment}
\usepackage{natbib}

\usepackage{caption}
\usepackage{subcaption}

\usepackage{mathtools}

\DeclarePairedDelimiter\floor{\lfloor}{\rfloor}

\title{On the importance of catalyst-adsorbate 3D interactions for relaxed energy predictions}

\author{
    Alvaro Carbonero \\
  Mila \\
  \texttt{alvaro.carbonero@mila.quebec} \\
  \AND
  Alexandre Duval \\
  Mila, CentraleSupélec \\
  \And  
  Victor Schmidt \\
  Mila, Université de Montréal \\
  \And 
  Santiago Miret \\
  Intel labs \\
  \And 
  Alex Hernandez-Garcia \\
  Mila, Université de Montréal
  \And 
  Yoshua Bengio \\
  Mila, Université de Montréal
  \And 
  David Rolnick \\
  Mila, McGill University \\
}

\begin{document}

\maketitle

\begin{abstract}
The use of machine learning for material property prediction and discovery has traditionally centered on graph neural networks that incorporate the geometric configuration of all atoms. However, in practice not all this information may be readily available, e.g.~when evaluating the potentially unknown binding of adsorbates to catalyst. In this paper, we investigate whether it is possible to predict a system's relaxed energy in the OC20 dataset while ignoring the relative position of the adsorbate with respect to the electro-catalyst. We consider SchNet, DimeNet++ and FAENet as base architectures and measure the impact of four modifications on model performance: removing edges in the input graph, pooling independent representations, not sharing the backbone weights and using an attention mechanism to propagate non-geometric relative information. 
We find that while removing binding site information impairs accuracy as expected, modified models are able to predict relaxed energies with remarkably decent MAE. Our work suggests future research directions in accelerated materials discovery where information on reactant configurations can be reduced or altogether omitted.
\end{abstract}


\section{Introduction}
\label{sec:introduction}

Materials discovery is a strong driver of innovation, unlocking new materials with tailored properties that benefit various domains including energy efficiency, transportation systems and electronics \citep{butler2018machine}. Yet, it faces significant hurdles. Characterizing new materials is computationally expensive \citep{oganov2019structure}, even when replacing lab experiments by quantum mechanical simulations like the Density Functional Theory (DFT) \citep{kohn1996density}. Besides, exploration of materials is hindered by the vastness of the search space, which encompasses myriad compositions, atomic arrangements and properties~\citep{pyzer2022accelerating}.

To overcome these challenges, researchers have turned to Machine Learning (ML) for accelerated materials discovery \citep{zhang2023survey, miret2022open, lee2023matsciml, song-etal-2023-matsci} for two primary reasons: First, ML holds the power to quickly model materials' properties (i.e., to evaluate candidates) using Geometric Graph Neural Networks (GNNs). Second, generative ML can automatically propose new and consistent material candidates.

In recent years, ML has emerged as a crucial tool for the discovery of electro-catalysts, which play a key role in promoting renewable energy processes and sustainable chemical production, including the production of ammonia for fertilizers and hydrogen \citep{zitnick2020introduction}. The Open Catalyst Project (OCP) \citep{chanussot2021open} has significantly contributed to this field by releasing an extensive dataset of 1,281,040 DFT relaxations of catalyst-adsorbate pairs, selected from a pool of meanigful candidates\footnote{The adsorbate refers to the molecule involved in the electrochemical reaction that is accelerated through the introduction of a catalyst material, represented by a semi-infinite periodic substructure.}. This dataset was specifically designed to train ML models to predict the relaxed energy of 3D adsorbate-catalyst (adslab) systems, a critical property that influences a catalyst’s activity and selectivity \citep{tran2018active}, or its effectiveness for a specific chemical reaction\footnote{In other words, it quantifies the extent to which the catalyst reduces the energy required for a chemical reaction to occur.}. OCP has facilitated significant advancements in catalysis discovery, with ML models increasingly bridging the performance gap with DFT simulations, while offering a speed advantage of several orders of magnitude \citep{musaelian2022learning, gasteiger2022gemnet, passaro2023reducing}.

While research has primarily focused on predicting material properties \citep{wieder2020compact}, it’s equally crucial to efficiently explore the vast search space of potential catalyst candidates \citep{wei2019machine, zhang2022machine}. This involves generating consistent candidates automatically, a challenge due to the intricate process of creating adslab samples. This process, further detailed in Appendix \ref{app:sec:adslab}, includes positioning an adsorbate molecule (e.g., $\textsc{H2O}$) with a catalyst in 3D space, meaning cutting a surface through the catalyst bulk, selecting the adsorbate’s spatial orientation, and sampling a plausible binding site \citep{chanussot2021open}. These steps, which are time-consuming and challenging to model, determine the input configuration of the adslab sample and significantly influence the relaxed energy prediction \citep{lan2022adsorbml}. An additional challenge is that this process and its relevance to the actual in-lab efficiency of a real material is not yet fully understood by chemists \citep{deshpande2020graph}.

In light of these challenges, this paper explores the possibility of predicting the relaxed energy of an adslab without co-locating the adsorbate and the catalyst in the same 3D space. This direction of study is valuable for several reasons:
(1) to better understand the role of adsorbate-catalyst geometry in determining the relaxed energy; (2) to reduce reliance on the exact input configuration, which often correlates with local energy minima, (3) to avoid the complexity and computational cost of determining a good input configuration (e.g., finding a binding site and adsorbate orientation). Our primary objective is quantify the loss in accuracy that occurs when the geometric relationship between inputs is unavailable.

To achieve this, we propose four modifications of existing GNN architectures, collectively referred to as \textit{Disconnected GNNs}. All four modifications enable the base architecture to make relaxed energy predictions without explicitly modeling geometric interactions between the adsorbate and the catalyst. We then evaluate the trade-off of omitting these interactions through experiments on the OC20 dataset and suggest future directions to overcome the limits associated with the input adslab configuration.


\section{Methods}
\label{sec:methods}

Our methods test the assumption that meaningful results can be obtained in predicting the relaxed energy even when the geometric interactions between the adsorbate and the catalyst are unknown. Specifically, we propose four \textit{Disconnected GNN} models. These models can leverage as their backbone any underlying GNN model that predict properties of 3D atomic systems using the general pipeline detailed under ``Refresher'' below, where we make targeted changes to the pipeline. In this paper, we investigate the effect of the proposed Disconnected GNN architecture on three backbone models: SchNet \citep{schutt2017schnet}, DimeNet++ \citep{klicpera2020fast} and FAENet \citep{duval2023faenet}.

\textbf{Refresher}. Recall that GNNs applied to 3D atomic systems mostly use the following pipeline. (1) \textbf{Graph creation}: construct a graph representation of the input point cloud systems (e.g.~adslab), represented with atomic numbers and 3D atom positions.
(2) \textbf{Embedding}: derive node/edge embeddings based on atomic numbers and geometric information (e.g. relative atom positions). (3) \textbf{Interaction blocks}: apply a fixed number of message passing layers~\citep{gilmer2017neural} to update the node and/or the edge embeddings, using geometric information and preserving data symmetries. (4) \textbf{Output block}: project final atom representations into scalar values (e.g.~atom-properties). For graph-level prediction, a global pooling (e.g.~sum) is performed to aggregate atom predictions.

\begin{figure}
    \centering
    \begin{subfigure}[b]{0.2\textwidth}
        \includegraphics[scale=0.75]{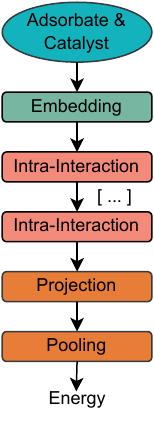}
        \caption{Baseline.}
        \label{baseline}
    \end{subfigure}
    \begin{subfigure}[b]{0.3\textwidth}
        \includegraphics[scale=0.75]{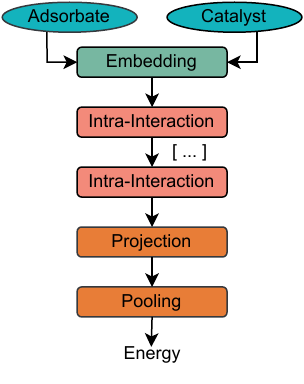}
        \caption{Disconnected baseline.}
        \label{disconnected}
    \end{subfigure}
    \vspace{0.2cm}
    \begin{subfigure}[b]{0.3\textwidth}
        \includegraphics[scale=0.75]{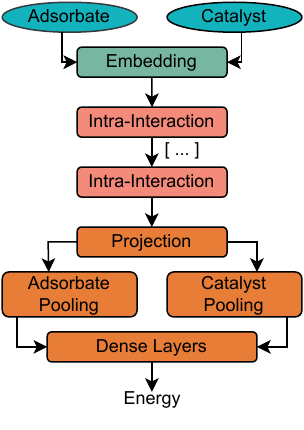}
        \caption{Independent pooling.}
        \label{dependent}
    \end{subfigure}
    \begin{subfigure}[b]{0.4\textwidth}
        \centering
        \includegraphics[scale=0.75]{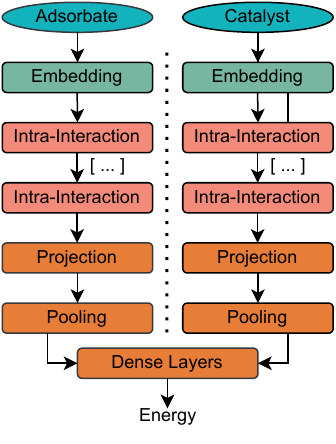}
        \caption{Independent Backbones.}
        \label{independent}
    \end{subfigure}
    \begin{subfigure}[b]{0.4\textwidth}
        \centering
        \includegraphics[scale=0.75]{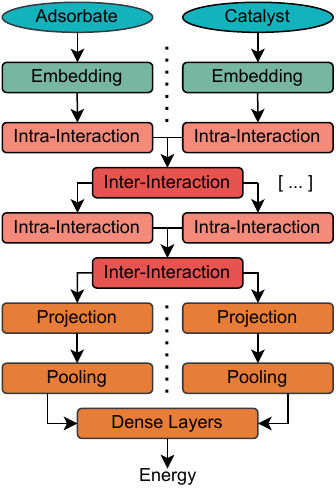}
        \caption{Attention.}
        \label{attention}
    \end{subfigure}
    \caption{Architectures are presented in order of increasing complexity. \ref{baseline} illustrates the standard property prediction pipeline for Geometric GNNs, with no modifications. In \ref{disconnected}, edges between parts are removed. In \ref{dependent}, the pooling step is performed separately on atom representations for the adsorbate and the catalyst. Resulting embeddings are concatenated and passed through an MLP. In \ref{independent}, each part has its own model with distinct weights. In \ref{attention}, an interaction block with artificial edges and weights allows node embeddings between both parts to communicate.}
    \label{architectures}
\end{figure}

\subsection{Disconnected baseline}
\label{subsec:disconnected-model}

The baseline Disconnected GNN model does not create any edges between the catalyst and the adsorbate in the graph creation step. By removing all such edges, no relative geometric information will pass from the adsorbate to the catalyst when modelling the system, since GNNs only propagate information through graph edges. Since this modification only relates to the input data representation, this change can be applied to any backbone GNN architecture, like FAENet, SchNet or DimeNet++ for instance. It is illustrated in Figure \ref{disconnected}.

Note that we still use the original OC20 dataset where the adsorbate/catalyst atom positions have been determined after selecting a binding site and an adsorbate orientation. But since GNNs use atom relative positions and not absolute atom positions to preserve translation equivariance, disconnecting the adsorbate-catalyst graph is enough to discard totally the input configuration. Processing such a disconnected graph is equivalent to considering both components independently. Associated results thus inform about the importance of the relative adslab geometric information to each model.

\subsection{Independent pooling}
\label{subsec:dependent-model}

This method builds upon the disconnected baseline defined above, also removing all edges between the adsorbate and the catalyst. As illustrated in Figure \ref{dependent}, there are three additional changes. First, the projection block now outputs a hidden representation $\mathbf{h}\in \mathbb{R}^{\floor*{H/2}}$ for each node instead of a scalar quantity, where $H$ is the dimension of the embedding. Second, the resulting node representations are pooled separately for each component, leading to distinct hidden representations for the adsorbate and the catalyst. Lastly, we concatenate these two representations and pass them to an additional dense layer to compute the energy of the system. This final MLP layer gives more expressive power to the model than the disconnected baseline because we explicitly model non-geometric interactions between the adsorbate and the catalyst.

\subsection{Independent backbones}
\label{subsec:independent-model}

This approach, illustrated in Figure \ref{independent}, builds upon independent pooling. The main difference is that catalysts and adsorbates are assigned distinct GNN models (hence ``independent”) instead of sharing backbone weights. 
The motivation is that the adsorbate and the catalyst have very different sizes and roles. Hence, it might be beneficial to have independent GNNs, where each one is trains for its particular input.
Similarly to independent pooling, both models are modified to produce a hidden representation of their respective parts. They are then concatenated and passed through an MLP.

\subsection{Attention}
\label{subsec:attention-model}

This model (see Figure \ref{attention}) builds upon the independent model by allowing the nodes of both graphs to communicate during the interaction layers.

We modify the graph creation process to produce a heterogeneous graph where we add the following weighted edges to previous models' edges.
For every catalyst node $i$ and adsorbate node $j$, we create the new edges $(i, j)$ and $(j, i)$ with weights $z$ and $-z$ respectively, where $z$ is the $z$-axis coordinate of $x_i$. Besides, in the embedding step, we pass the edge weights through an RBF layer and a linear layer to increase their dimension. Notice that this model can run without locating the adsorbate and catalyst in the same 3D plane since no information regarding the adsorbate's location (e.g. distance, relative position) is given.
The motivation behind these weighted edges is to make the model aware of the closeness of a node to the catalyst's surface, while respecting the restrictions of the setup. As noted in \citep{duval2022phast}, fixed atoms not near the surface of the catalyst (tag 0 nodes) can be removed without hurting efficiency, indicating that closeness to surface plays a factor. 

We now describe how this graph is used. Let $F$ denote a ``normal'' intra-interaction layer, let $h^l_{\text{ads}}$ and $h^l_{\text{cat}}$ denote the node embeddings of the adsorbate and the catalyst at interaction layer $l$, and let $e_{\text{ads}}$ and $e_{\text{cat}}$ stand for the edges of the adsorbate and catalyst, respectively. To produce $h^{l+1}_{\text{ads}}$ and $h^{l+1}_{\text{cat}}$, we first calculate $h'_{\text{ads}} = F(h^l_{\text{ads}}, e_{\text{ads}})$ and $h'_{\text{cat}} = F(h_{\text{cat}}, e_{\text{cat}})$. Subsequently, we calculate $h' = \text{GATConv}(\text{Concat}(h'_{\text{ads}}, h'_{\text{cat}}), e')$ where $e'$ is the set of new weighted edges and GATConv is a convolution layer as described in \citep{velivckovic2017graph}. Finally,
\[
h^{l+1}_{\text{ads}}, h^{l+1}_{\text{cat}} = \text{Norm}(h'_{\text{ads}} + h^l_{\text{ads}}), \text{Norm}(h'_{\text{cat}} + h^l_{\text{cat}}).
\]


\section{Evaluation}
\label{sec:eval}

\textbf{Dataset}: We use the OC20 IS2RE~\citep{zitnick2020introduction} dataset, which involves the direct prediction of the relaxed adsorption energy from the initial atomic structure, i.e. a graph regression task requiring E(3)-invariance. OC20 comes with $450K$ training samples and a predefined train/val/test split. We evaluate models on the four distinct splits of the validation set ($\sim25K$ samples each): In Domain (ID), Out of Domain Adsorbates (OOD-ads), Out of Domain catalysts (OOD-cat), and Out of Domain Adsorbates and catalysts (OOD-both).

\textbf{Metrics}: We measure accuracy on each validation split through the energy MAE. Running time is measured with the throughput at inference time, i.e. the average number of samples per second that a model can process in its forward pass, on similar GPU types. 

\textbf{Baselines}: 
We compare base results of the original models (considering 3D adsorbate-catalyst interactions) with the four disconnected architectures detailed in Section \ref{sec:methods}. Note that we apply PhAST components \citep{duval2022phast} on all models to improve performance and inference time.

\begin{figure}
    \centering
    \includegraphics[width=\textwidth]{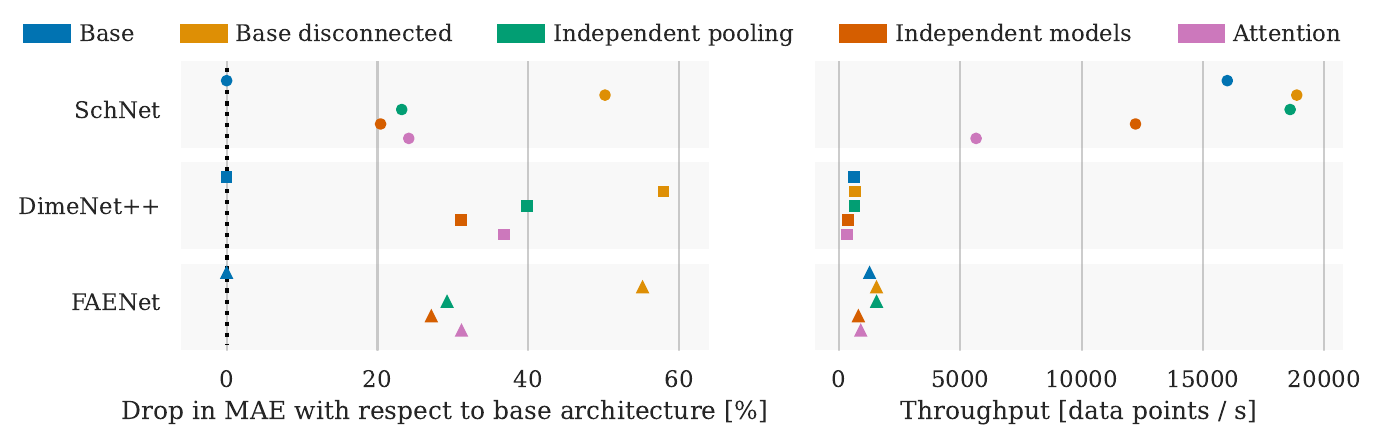}
    \caption{Comparison of MAE performance drops across our four proposed modifications to three base architectures (left), and their associated impact on model throughput (right). We can see the the disconnected baseline consistently achieves the best performance. In terms of throughput, it is on par with other modifications for DimeNet++ and FAENet, while it is slower than all other modifications for SchNet (except for the Attention modification).}
    \label{fig:results}
\end{figure}
 
\textbf{Results}. From Figure~\ref{fig:results}, we see that the independent model architecture performs best among all proposed disconnected GNNs. However, it presents a performance drop in MAE of 20\%, 31\%, and 28\% with respect to baseline architectures, for Schnet, DimeNet++ and FAENet respectively. A complete numerical table of results is reported in Table~\ref{tab:results}.

Performance is significantly worse for disconnected GNNs than for GNNs with access to the adslab configuration; however, \textbf{these results are promising in several ways}. 

Firstly, a drop in performance was expected since we neglect important information by omitting to modelise the adsorbate-catalayst 3D interaction. The important take-away is that our disconnected GNN manages to produce decent energy predictions while being agnostic to the adslab's input configuration. Future work will probably manage to reduce further this performance gap. Note that we discuss some axes of improvements in Appendix \ref{app:sec:improvements}.

Secondly, the OC20 dataset often contains multiple input configuration samples for the same catalyst-adsorbate pair, each with a different relaxed energy target. This cannot be captured by our disconnected GNN, leading to multiple targets for the same sample, which can impair training dynamics. Furthermore, when evaluating on a dataset that does not present this feature, e.g.~the 10k split of OC20 IS2RE, disconnected GNNs match (or outperform) baseline models, which is very promising. We present and briefly discuss these results in Appendix \ref{app:10k}.

The \textsl{OC20-Dense} dataset \citep{lan2022adsorbml} was recently released, containing multiple input configurations (e.g. binding site, adsorbate orientation) for each adsorbate-catalyst pair along with the associated relaxed energy. Building upon this dataset, we will be able to train disconnected GNN models to predict the minimum relaxed energy of each adslab over all possible input configurations. The obvious benefit of such approach is that it marginalises over adslab binding sites and adsorbate orientations, avoiding the need to explore all configurations to actually find the relaxed energy global minima of this adsorbate-catalyst pair, which is expected to be the main reaction driver in real life experiments \citep{lan2022adsorbml}.

\section{Conclusion}

In this paper, we have explored predicting an adsorbate-catalyst system’s relaxed energy without co-locating them in 3D space. We proposed four disconnected models built on top of existing Geometric GNNs architectures, and showed through experiments that this task was possible with a manageable performance loss. Looking forward, we see the potential of using such disconnected models to find the global energy minima of adslab systems while circumventing the need to consider all possible input adslab configurations. This sounds particularly promising for generative methods and should contribute to the acceleration of catalysis discovery.

\bibliography{references}
\bibliographystyle{plainnat}

\appendix

\section{Adslab generation}
\label{app:sec:adslab}

This section summarises the adslab creation process of the OC20 dataset. Refer to the original paper \citep{chanussot2021open} for more details. 

\textbf{Adsorbate and catalyst surface selection}. The first step is to select the adsorbate and the catalyst that will compose the adslab. For the adsorbate, it extremely simple: it is sampled randomly from a set of 82 molecules that are chosen for their utility to renewable energy applications. 
For the catalyst surface, the process is divided in three stages. First, we choose the number of distinct chemical elements composing it. It can be a unary material (5\% chance), a binary material (65\%) or a ternary material (30\%). These elements are chosen from a set of 55 elements comprising reactive nonmetals, alkali metals, transition metals, etc. 
Next, a stable bulk material is randomly selected from the 11K samples of the Materials Project \citep{jain2013materials} with the number of elements chosen in the first step. Lastly, all symmetrically distinct surfaces\footnote{a surface is a cut through the bulk operated using Miller indices} from the material with Miller indices \citep{ashcroft2022solid} less than or equal to 2 are enumerated, including possibilities for different absolute positions of surface plane. From this list of surfaces, one is randomly selected.

\textbf{Input adslab configuration}. The objective of the second step is to place the adsorbate and the catalyst surface in the same 3D plane. Using pymatgen’s Voronoi tesselation algorithm \citep{ong2013python} on surface atoms and adsorbate possible binding sites, Catkit \citep{boes2019graph} enumerates a list of symmetrically distinct adsorption sites along with suggested per-site adsorbate orientations. From this list, an adsorption configuration is randomly selected, yielding the initial adslab structure. This initial structure is then relaxed using Density Functional Theory simulations, performed using the Vienna Ab Initio simulation Package (VASP) \citep{kresse1994ab}.

\section{Full results table}

\begin{table*}[t]
\centering
\resizebox{\textwidth}{!}{\begin{tabular}{lccccc|c}
\cmidrule[1.3pt]{1-7}
\textbf{Baseline / MAE} & ID & OOD-ad & OOD-cat & OOD-both & Average & Throughput (samples/s) \\
\cmidrule[.5pt]{1-7}
\rowcolor{cycle2!8} \textsl{SchNet}     & $0.631$ & $0.687$ & $0.625$ & $0.626$ & $\pmb{0.642}$ & $16001 \pm 2264$ \\
\textsl{Base disconnected-SchNet}    & $0.940$ & $1.037$ & $0.932$ & $0.947$ & $0.964$  & \pmb{$18863 \pm 3034$}\\
\textsl{Independent pooling-SchNet}    & $0.766$ & $0.847$ & $0.762$ & $0.789$ & $0.791$  & \underline{$18590 \pm 2230$}\\
\textsl{Independent models-SchNet}    & $0.761$ & $0.805$ & $0.764$ & $0.764$ & $\underline{0.773}$  & $12220 \pm 1593$\\
\textsl{Attention-SchNet}    & $0.779$ & $0.852$ & $0.762$ & $0.797$ & $0.797$  & $5652 \pm 876$\\
\cmidrule[1.3pt]{1-7}
\rowcolor{cycle2!8} \textsl{DimeNet++}    & $0.577$ & $0.693$ & $0.568$ & $0.621$ & $\textbf{0.615}$  & $631 \pm 125$\\
\textsl{Base disconnected-DimeNet++}    & $0.922$ & $1.059$ & $0.914$ & $0.989$ & $0.971$  & \pmb{$667 \pm 143$}\\
\textsl{Independent pooling-DimeNet++}    & $0.781$ & $0.0.992$ & $0.767$ & $0.901$ & $0.860$  & \underline{$650 \pm 139$}\\
\textsl{Independent models-DimeNet++}    & $0.767$ & $0.884$ & $0.754$ & $0.818$ & $\underline{0.806}$  & $374 \pm 68$\\
\textsl{Attention-DimeNet++}    & $0.777$ & $0.952$ & $0.763$ & $0.870$ & $0.841$  & $329 \pm 57$\\
\cmidrule[.5pt]{1-7}
\rowcolor{cycle2!8} \textsl{FAENet}    & $0.554$ & $0.623$ & $0.546$ & $0.578$ & $\textbf{0.575}$  & $1263 \pm 766$\\
\textsl{Base disconnected-FAENet}    & $0.884$ & $0.937$ & $0.878$ & $0.870$ & $0.892$  & $1549 \pm 1027$\\
\textsl{Independent pooling-FAENet}    & $0.698$ & $0.813$ & $0.697$ & $0.766$ & $0.743$  & \underline{$1553 \pm 1049$}\\
    \textsl{Independent models-FAENet}    & $0.690$ & $0.782$ & $0.689$ & $0.760$ & $\underline{0.731}$  & \pmb{$803 \pm 327$}\\
\textsl{Attention-FAENet}    & $0.732$ & $0.814$ & $0.715$ & $0.756$ & $0.754$  & $902 \pm 565$\\
\cmidrule[.5pt]{1-7}
\end{tabular}}
\caption{MAE and inference time for various GNNs and their disconnected counterpart on \textsl{OC20} IS2RE, averaged over 3 runs. \textit{Average} MAE is computed over all validation splits. Best results are shown in bold, second best underlined. Overall, disconnected models show a lower accuracy than baselines but they are able to predict relaxed energy relatively well despite omitting geometric interaction between parts in the adslab.} 
\label{tab:results}
\end{table*}

\section{Axes of improvement}
\label{app:sec:improvements}

There are two main ways in which Disconnected GNN models can be further improved. The first one is to improve upon dense layers as a way to combine the final hidden representations of the adsorbate and the catalyst. The second one is to find a way to effectively allow the nodes of the adsorbate and catalyst to communicate, and to do this while ommitting the relative position and orientation between the adsorbate and the catalyst. This is attempted in the attention models through the inter-interaction layers. Given that they do not consistently outperform the independent backbone models, it is likely that the way in which attention models use GAT convolution layers hinders the model's prediction. We still hope that this can be fixed since the attention models can outperform baseline models when trained on the 10k dataset, as we described in Appendix \ref{app:10k}.

\section{Results in the 10k dataset}
\label{app:10k}

Only 89\% of the OC20 IS2RE dataset consists of adsorbate-catalyst systems which are unique up to the combination of adsorbate id, bulk id, and bulk cell. That is, to Disconnected GNNs, 11\% of the dataset consists of inputs with multiple targets due to distinct binding sites, which impairs training dynamics. When compared to the OC20 10k IS2RE dataset, every single adsorbate-catalyst system has a unique combination of adsorbate id, bulk id, and bulk cell. Thus, evaluating disconnected GNNs on the 10k dataset is a useful way to assess their performance on a suitable data tailored to them. This is corroborated by the results in Table \ref{tab:results10k}.

\begin{table*}[t]
\centering
\resizebox{\textwidth}{!}{\begin{tabular}{lccccc|c}
\cmidrule[1.3pt]{1-7}
\textbf{Baseline 10k/ MAE} & ID & OOD-ad & OOD-cat & OOD-both & Average & Throughput (samples/s) \\
\cmidrule[.5pt]{1-7}
\rowcolor{cycle2!8} \textsl{SchNet}     & $1.151$ & $1.245$ & $1.104$ & $1.213$ & $1.179$ & $18602 \pm 4373$ \\
\textsl{Base disconnected-SchNet}    & $1.246$ & $1.236$ & $1.214$ & $1.144$ & $1.210$  & $25487 \pm 3693$\\
\textsl{Independent pooling-SchNet}    & $0.962$ & $1.038$ & $0.935$ & $0.961$ & \underline{$0.974$}  & $24832 \pm 3041$\\
\textsl{Independent models-SchNet}    & $0.976$ & $0.985$ & $0.946$ & $0.935$ & $\pmb{0.961}$  & $\underline{16501 \pm 3283}$\\
\textsl{Attention-SchNet}    & $0.979$ & $1.072$ & $0.961$ & $0.968$ & $0.995$  & $\pmb{7952 \pm 1261}$\\
\cmidrule[1.3pt]{1-7}
\rowcolor{cycle2!8} \textsl{DimeNet++}    & $0.851$ & $0.969$ & $0.802$ & $0.875$ & $\textbf{0.874}$  & $623 \pm 123$\\
\textsl{Base disconnected-DimeNet++}    & $1.044$ & $1.078$ & $1.026$ & $0.971$ & $1.023$  & $649 \pm 140$\\
\textsl{Independent pooling-DimeNet++}    & $0.908$ & $1.022$ & $0.888$ & $0.928$ & $0.937$  & $752 \pm 114$\\
\textsl{Independent models-DimeNet++}    & $0.889$ & $0.971$ & $0.876$ & $0.883$ & $\underline{0.905}$  & $\underline{366 \pm 71}$\\
\textsl{Attention-DimeNet++}    & $0.889$ & $1.017$ & $0.865$ & $0.915$ & $0.921$  & $\pmb{330 \pm 60}$\\
\cmidrule[.5pt]{1-7}
\rowcolor{cycle2!8} \textsl{FAENet}    & $1.003$ & $1.017$ & $0.992$ & $1.004$ & $1.004$  & $\underline{1417 \pm 792}$\\
\textsl{Base disconnected-FAENet}    & $1.104$ & $1.144$ & $1.118$ & $1.061$ & $1.107$  & $1492 \pm 1085$\\
\textsl{Independent pooling-FAENet}    & $1.022$ & $1.009$ & $0.999$ & $0.938$ & $0.992$  & $1626 \pm 1123$\\
\textsl{Independent models-FAENet}    & $0.891$ & $1.020$ & $0.871$ & $1.040$ & $\underline{0.955}$  & $1089 \pm 737$\\
\textsl{Attention-FAENet}    & $0.894$ & $0.924$ & $0.863$ & $0.872$ & \pmb{$0.888$}  & $\pmb{916 \pm 567}$\\
\cmidrule[.5pt]{1-7}
\end{tabular}}
\caption{MAE and inference time for various GNNs and their disconnected counterpart on \textsl{OC20} 10k IS2RE, averaged over 3 runs. \textit{Average} MAE is computed over all validation splits. Disconnected models all show lower accuracy. In every model, every subsurface catalyst atom (tag 0 atoms) is removed as suggested in \citep{duval2022phast} to improve inference time.} 
\label{tab:results10k}
\end{table*}

From Table \ref{tab:results10k}, we see that independent models perform on par or better (for FAENet) than baseline models when trained on the 10k dataset. In particular, we see a performance improvement of 18\%, -4\%, and 5\% for SchNet, DimeNet++, and FAENet respectively. Different than in the all dataset, Attention-FAENet outperforms Independent models-FAENet by 7.5\%. These results are a sign that Disconnected GNNs can perform well when tested against the data for which they were designed for. The OC20-Dense dataset \citep{lan2022adsorbml} is thus an exciting development as it will allow for the creation of the datasets that these models need.

\end{document}